\documentclass[11pt]{article}
\usepackage{syntaxfest2021}
\usepackage{natbib}
\usepackage{times}
\usepackage{url}
\usepackage{latexsym}
\usepackage[T1]{fontenc}
\usepackage[utf8]{inputenc}
\usepackage[table,xcdraw]{xcolor}
\usepackage{multirow}
\usepackage{booktabs}
\usepackage{todonotes}

\usepackage[breaklinks,hidelinks]{hyperref}

\syntaxfestfinalcopy 

\title{Parsing with Pretrained Language Models, Multiple Datasets, and Dataset Embeddings}

\author{Rob van der Goot \\
  IT University of Copenhagen\\
  {\tt robv@itu.dk} \\\And
  Miryam de Lhoneux \\
  Uppsala University \\KU Leuven \\University of Copenhagen\\
  {\tt ml@di.ku.dk} \\}

\date{}

\begin{document}
\maketitle
\begin{abstract}
With an increase of dataset availability, the potential for learning from a variety of data sources has increased. One particular method to improve learning from multiple data sources is to embed the data source during training. This allows the model to learn generalizable features as well as distinguishing features between datasets. However, these dataset embeddings have mostly been used before contextualized transformer-based embeddings were introduced in the field of Natural Language Processing. In this work, we compare two methods to embed datasets in a transformer-based multilingual dependency parser, and perform an extensive evaluation. We show that: 1) embedding the dataset is still beneficial with these models 2) performance increases are highest when embedding the dataset at the encoder level 3) unsurprisingly, we confirm that performance increases are highest for small datasets and datasets with a low baseline score. 4) we show that training on the combination of all datasets performs similarly to designing smaller clusters based on language-relatedness.\footnote{Code available at: \url{https://bitbucket.org/robvanderg/dataembs2}, our implementations will also be included in the next MaChAMp release (v0.3)}
\end{abstract}

\section{Introduction}
\label{sec:intro}

Many studies have shown the benefits of training dependency parsers jointly for multiple treebanks, either within the same language~\citep{stymne-etal-2018-parser} or across different languages~\citep[multilingual models;][]{ammar-etal-2016-many, vilares-etal-2016-one,smith-etal-2018-82}, which makes it possible to transfer knowledge between treebanks.
This has been enabled by the development of treebanks in multiple languages annotated according to the same guidelines which have been released by the Universal Dependencies \citep[UD;][]{nivre-etal-2020-universal} project. In this context, a method which has been shown to be effective is the use of embeddings that represent each individual treebank. This is referred to as a \emph{language embedding}~\citep{ammar-etal-2016-many} or a \emph{treebank embedding}~\citep{smith-etal-2018-82}, we opt for the more general term: \emph{dataset embedding}.
The main intuition behind these dataset embeddings is that they allow the model to learn useful commonalities between different datasets, while still encoding dataset specific knowledge.

In the last few years, large multilingual pretrained language models (LMs) pretrained on the task of language modelling, such as mBERT \citep{devlin-etal-2019-bert} or XLM-R \citep{conneau-etal-2020-unsupervised}, have made it possible to train high performing multilingual models for many different NLP tasks, including dependency parsing \citep{kondratyuk-straka-2019-75}. 
These models have shown surprising cross-lingual abilities in spite of not getting any cross-lingual supervision. \citet{wu-dredze-2019-beto}, for example, fine-tuned mBERT on English data for five different tasks and applied it to different languages with high accuracy, relative to the state-of-the-art for these tasks.
These large pretrained multilingual LMs are now widely used in multilingual NLP.

Dataset embeddings have so far only been evaluated in the context of multilingual dependency parsing without such large pretrained multilingual LMs. Given the large gains in accuracy that have been obtained by using them, and given that they seem to be learning to do cross-lingual transfer without cross-lingual supervision, it is unclear whether or not datataset embeddings can still be useful in this context. This is the question we ask in this paper. Our main research question can be formulated as follows:

\begin{itemize}
  \item[\textbf{RQ}] Is the information learned by dataset embeddings complementary to the information learned by large multilingual LM-based parsers?
\end{itemize}

\section{Background}

\citet{vilares-etal-2016-one} were among the first to exploit UD treebanks to train models for multiple languages. They trained parsing models on 100 pairs of languages by simply concatenating the treebank pairs. They observed that most models obtained comparable accuracy to the monolingual baseline and some outperformed it. This shows that even in its simplest form, multilingual training can be beneficial. \citet{ammar-etal-2016-many} build a more complex model to train a parser for eight languages. They introduce a vector representation of each language which is used as a feature, concatenated to the word representations of each word in the sentence.

\citet{smith-etal-2018-82} used this idea in the context of the CoNLL 2018 shared task \citep{zeman-etal-2018-conll} where the task was to parse test sets in 82 languages. \citet{smith-etal-2018-82} found that training models on clusters of related languages using dataset embeddings led to a substantial accuracy gain over training individual models for each language. 

\citet{stymne-etal-2018-parser} further exploited this dataset embedding method in the monolingual context, using heterogeneous treebanks. They compared this method to several methods including simply concatenating the treebanks to learn a unique parser for all treebanks and found the dataset embedding method to be superior to all other methods.

\citet{wagner-etal-2020-treebank} investigated whether dataset embeddings can be useful in an out-of-domain scenario where the treebank of the target data is unknown. They found that it is possible to predict dataset embeddings for such target treebanks, making the method useful in this scenario.

\citet{kondratyuk-straka-2019-75} trained a single model for the 75 languages available in UD at the time by concatenating all treebanks and using a large pretrained LM. They found that this did not hurt accuracy compared to training monolingual models, and it even improved accuracy for some languages. This type of model has become standard and has been used in many studies. They did not make use of dataset embeddings in this setup. 

Dataset embeddings have mostly been used with BiLSTM parsers. This may partially be due to the fact that they were concatenated to the word embedding before passing it into the encoder, which is non-trivial in LM-based setups where the word embedding and encoding size is fixed. To the best of our knowledge, the only attempt to use dataset embeddings in combination with a LM-based parser was from~\citet{van-der-goot-etal-2021-massive}. They use a large pretrained multilingual LM as encoder, and concatenate the dataset embeddings to the word embedding before decoding. They show that this leads to improved performance if the task is the same and the languages/domains of the datasets differ. For a setup where different tasks are combined via multi-task learning (i.e. GLUE), dataset embeddings helped to decrease the performance gap compared to single-task models. 

\paragraph*{Contributions}

In this work, 1) we introduce a method to incorporate dataset embeddings also in the encoder in LM-based parsers; 2) we test whether or not dataset embeddings are useful when used in combination with large pretrained multilingual LMs, using both our newly proposed method as well as the existing approach; 3) we compare the effectiveness of dataset embedding when training on small clusters of datasets as well as training on all considered datasets simultaneously.

\section{Methodology}
\subsection{Methods}
\begin{figure}
    \centering
    \definecolor{armygreen}{rgb}{0.29, 0.33, 0.13}
\definecolor{brickred}{rgb}{0.8, 0.25, 0.33}
\definecolor{darksalmon}{rgb}{0.91, 0.59, 0.48}
\definecolor{deeppeach}{rgb}{1.0, 0.8, 0.64}
\definecolor{deepchampagne}{rgb}{0.98, 0.84, 0.65}
\definecolor{darkgreen}{rgb}{0.0, 0.2, 0.13}
\definecolor{airforceblue}{rgb}{0.36, 0.54, 0.66}

\def\embed#1#2#3
{
\begin{scope}
\def\temp{#3}\ifx\temp\empty
\else
    \node (text) [line width=.05cm, draw, deepchampagne, minimum width=3.25cm, minimum height=.5cm] at (#1, #2+.5) {\color{black}#3};
\fi

\node (box) [rectangle, fill,line width=.05cm, draw, opacity=.5, deepchampagne, minimum width=3.25cm, minimum height=.5cm] at (#1, #2) {};
\newcount\wordPiece
\wordPiece=0
\loop
    \node (circle) [circle, draw, minimum width=.2cm, minimum height=.2cm] at (#1 -1.35 + \the\wordPiece * .45, #2) {};
    \advance \wordPiece +1
\ifnum \wordPiece<7
\repeat
\end{scope}
}

\def\dataembed#1#2#3
{
\begin{scope}
\node (text) [line width=.05cm, draw, dashed, airforceblue, minimum width=3.25cm, minimum height=.5cm] at (#1, #2+.5) {\color{black}#3};
\node (box) [rectangle, fill,dashed, line width=.05cm, draw, opacity=.5, airforceblue, minimum width=3.25cm, minimum height=.5cm] at (#1, #2) {};
\newcount\wordPiece
\wordPiece=0
\loop
    \node (circle) [circle, draw, minimum width=.2cm, minimum height=.2cm] at (#1 -1.35 + \the\wordPiece * .45, #2) {};
    \advance \wordPiece +1
\ifnum \wordPiece<7
\repeat
\end{scope}
}

\begin{tikzpicture}
    \path[use as bounding box] (-.5,.2) rectangle (15.5,6.25);

\node (input) [minimum height=1cm, text height=1,text depth=.25cm] at (0,  0) {Input};

\dataembed{3}{5.4}{Dataset embed.}
\embed{3}{4.05}{Position embed.}
\embed{3}{2.7}{Segment embed.}
\embed{3}{1.35}{Word embeddings}
\embed{3}{.25}{}
\node [minimum height=1cm, text height=1,text depth=.25cm] at (3,  2.1) {+};
\node [minimum height=1cm, text height=1,text depth=.25cm] at (3,  3.45) {+};
\node [minimum height=1cm, text height=1,text depth=.25cm] at (3,  4.8) {+};
\node [minimum height=1cm, text height=1,text depth=.25cm] at (3,  .6) {=};

\draw [->] ([yshift=.25cm]input.east) -- (box.west);

\node (transformer) [minimum height=1cm, text height=1,text depth=.25cm] at (6,  0) {Transformer};
\node (encoder) [minimum height=1cm, text height=1,text depth=.25cm] at (6,  -.5) {(encoder)};

\draw [->] (box.east) -- ([yshift=.25cm]transformer.west);

\embed{9}{.25}{}
\draw [->] ([yshift=.25cm]transformer.east) -- (box.west);

\node (box2) [rectangle, fill,dashed, line width=.05cm, draw, opacity=.5, brickred, minimum width=1.5cm, minimum height=.5cm] at (11.3, .25) {};
    \node (circle) [circle, draw, minimum width=.2cm, minimum height=.2cm] at (10.8, .25) {};
    \node (circle) [circle, draw, minimum width=.2cm, minimum height=.2cm] at (11.25, .25) {};
    \node (circle) [circle, draw, minimum width=.2cm, minimum height=.2cm] at (11.7, .25) {};
\node (box) [rectangle, dashed, line width=.05cm, draw, opacity=.5, brickred, minimum width=1.5cm, minimum height=1cm] at (11.3, 1) {};
\node () [brickred, minimum width=1.5cm, minimum height=1cm] at (11.3, 1.25) {\color{black}Dataset};
\node () [brickred, minimum width=1.5cm, minimum height=1cm] at (11.3, .75) {\color{black}embed.};

\node (datasetID) [brickred, minimum width=1.5cm, minimum height=1cm] at (9, 4) {\color{black}Dataset\_id=X};

\draw [->, dashed] (datasetID.west) -- ([yshift=1.6cm, xshift=-2.75cm]datasetID.west);
\draw [->, dashed] ([yshift=.2cm]datasetID.south) -- ([yshift=.2cm]box.north);

\node (decoder) [minimum height=1cm, text height=1,text depth=.25cm] at (13.5,  0) {Decoder};

\draw [->] (box2) -- ([yshift=.25cm]decoder.west);

\node (output) [minimum height=.5cm, text height=.5,text depth=.25cm] at (15,  1) {Output};

\draw [->,] (decoder) -- ([yshift=.2cm, xshift=-.1cm]output.south);

\end{tikzpicture}
    \caption{Visualization of the two models to integrate dataset embeddings into transformer based parsers. The dataset embeddings for \textsc{decoder} is displayed with the red box (dashed on the right), and \textsc{encoder} is displayed in the blue box (dashed on the left).}
    \label{fig:models}
\end{figure}
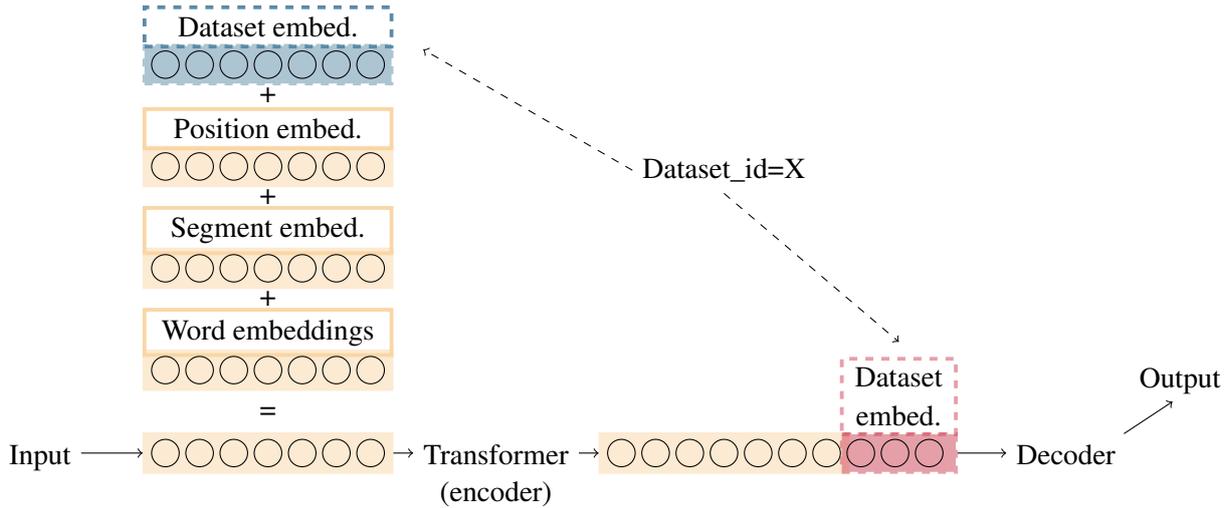

In most previous implementations of dataset embeddings, the dataset embedding is concatenated to the word embedding before it is passed into the encoder. When using the language model as encoder, as is now commonly done with BERT-like embeddings, this is impossible, as the word embedding is expected to be of a fixed size. For this reason, we experiment with two alternative setups: concatenating the embedding after the encoding, and summing it to the word embedding.
These two approaches are illustrated in Figure~\ref{fig:models}, and are explained in detail in the following two paragraphs. Note that these approaches can also be used simultaneously, which constitutes our third setup (\textsc{both}).

We use the deep biaffine parser~\citep{dozat2016deep} implementation of MaChAmp~\citep{van-der-goot-etal-2021-massive} as a framework for evaluating our models.
MaChAmp already includes an implementation of dataset embeddings on the decoder level.  In this implementation
the output of the LM for each wordpiece is concatenated to the dataset embedding before it is passed on to the decoder. We refer to this approach as \textsc{decoder}. In this setup, we choose to use dataset embeddings of size 12, based on previous work~\citep{ammar-etal-2016-many,smith-etal-2018-82}. This embedding is then concatenated to the output of the transformer, meaning that the input size of the decoder will be the size of the original output + 12.

The second approach incorporates the dataset embeddings in the LM parameters. 
In most transformer-based LM implementations, multiple embeddings are summed before the transformer layers to represent the input for each wordpiece. In the BERT model for example, these are the token embeddings, segment embeddings and position embeddings. We supplement these by also summing a dataset embedding to this input. In this way, the dataset embeddings can be taken into account throughout the transformer layers. We refer to this approach as \textsc{encoder}. In this setup, we choose to match the dimension size of the embeddings at the token level, to avoid having smaller embeddings summing only to an arbitrary subset of the weights.

\subsection{Experimental Setup}
We essentially reproduce the experiments from \citet{smith-etal-2018-82} using large pretrained multilingual LMs but do a more large-scale evaluation of the method, by testing more settings, comparing to more baselines and doing more extensive analysis. 
 More specifically, we use the clusters from~\citet{smith-etal-2018-82}, but use the updated versions of the treebanks (UD v2.8), and implement all models using MaChAmp, the library by \citet{van-der-goot-etal-2021-massive}. We use all default hyperparameters, including the mBERT embeddings~\citep{devlin-etal-2019-bert} which is used during the original tuning of MaChAmp~\cite{van-der-goot-etal-2021-massive}. All reported results are the average over 5 runs with different random seeds for MaChAmp. 

To avoid overfitting on the development or test set, we follow~\citet{van-der-goot-2021-need}, and compare our models on the development split while using a tune split for model picking. We will confirm our main findings on the test data in Section~\ref{sec:testData}. We use the updated splitting strategy proposed by~\citet{van-der-goot-2021-need}: for datasets with less then 3,000 sentences, we use 50\% for training, 25\% for tune, and 25\% for dev, for larger datasets we use 750 sentences for dev and tune, and the rest for train. We limit the size of each dataset to 20,000.

We compare models with dataset embeddings to baselines where we use the exact same parser (Figure~\ref{fig:models}), but without enabling any dataset emebddings. We use two training setups for our baselines: 1) monolingual models (\textsc{mono}) and 2) models where all treebanks are concatenated (\textsc{concat}). This is in contrast to \citet{smith-etal-2018-82} who only compared to a monolingual baseline.
We test this on the 59 test sets that are part of a cluster.\footnote{\citet{smith-etal-2018-82} trained mono-treebank models for the remaining test sets} Furthermore, we also explore what happens if we train one parser on all the datasets simultaneously, similar to Udify~\citep{kondratyuk-straka-2019-75}, who do not use dataset embeddings in their setup.

\section{Evaluation}

\subsection{Results}

\begin{table}[h!]
\centering
\setlength{\tabcolsep}{3pt}
\renewcommand{\arraystretch}{.7}
\resizebox{.92\textwidth}{!}{
\begin{tabular}{l l |c |cccc |cccc }
\toprule
& & & \multicolumn{4}{c|}{Trained on clusters} & \multicolumn{4}{c}{Trained on all}\\
\midrule
Cluster & Treebank & \textsc{mono} & \textsc{concat} & \textsc{decoder} & \textsc{encoder} & \textsc{both} & \textsc{concat} & \textsc{decoder} & \textsc{encoder} & \textsc{both} \\
\midrule 
af-de-nl & af\_afribooms & 80.63 & 82.12 & 82.62 & 81.17 & 80.93 & 82.98 & \textbf{83.53} & 82.62 & 82.37 \\
 & nl\_alpino & 92.75 & 93.01 & 92.97 & 92.50 & \textbf{93.15} & 92.65 & 92.59 & 93.08 & 92.76 \\
 & nl\_lassysmall & 87.25 & 89.71 & 89.91 & 89.59 & 89.69 & 89.45 & \textbf{90.29} & 90.11 & 90.05 \\
 & de\_gsd & 87.02 & 87.60 & 87.52 & 87.09 & 87.31 & 87.63 & \textbf{87.63} & 87.25 & 87.53 \\
\midrule 
e-sla & ru\_syntagrus & \textbf{94.75} & 94.74 & 94.56 & 94.60 & 94.67 & 94.50 & 94.45 & 94.51 & 94.52 \\
 & ru\_taiga & 74.93 & 74.54 & 75.11 & 75.75 & 76.21 & 75.43 & 75.95 & \textbf{76.46} & 76.45 \\
 & uk\_iu & 88.56 & 89.96 & 89.97 & 89.73 & 89.56 & 90.28 & \textbf{90.67} & 90.58 & 90.26 \\
\midrule 
en & en\_ewt & 89.34 & 88.90 & 89.06 & \textbf{89.67} & 89.12 & 88.49 & 88.16 & 88.86 & 88.94 \\
 & en\_gum & 90.38 & 89.06 & 90.17 & 90.74 & \textbf{90.91} & 88.64 & 89.65 & 90.53 & 90.71 \\
 & en\_lines & 86.57 & 84.72 & 84.77 & 87.00 & 87.38 & 84.70 & 85.08 & 87.41 & \textbf{87.42} \\
\midrule 
es-ca & ca\_ancora & 93.33 & 93.66 & 93.56 & 93.42 & \textbf{93.76} & 93.45 & 93.43 & 93.46 & 93.44 \\
 & es\_ancora & 92.99 & 93.08 & \textbf{93.38} & 93.30 & 93.30 & 93.35 & 93.28 & 93.24 & 93.16 \\
\midrule 
finno & et\_edt & \textbf{85.61} & 84.66 & 85.29 & 85.11 & 85.33 & 85.17 & 84.98 & 85.58 & 85.58 \\
 & fi\_ftb & 89.03 & 82.22 & 89.53 & \textbf{90.14} & 89.87 & 80.15 & 89.01 & 89.06 & 89.38 \\
 & fi\_tdt & 88.55 & 82.59 & 88.99 & 88.90 & 89.35 & 84.24 & 89.13 & \textbf{89.44} & 89.39 \\
 & sme\_giella$^*$ & 49.86 & 62.35 & 62.74 & 62.96 & 64.71 & 59.55 & 59.55 & 63.95 & \textbf{65.28} \\
\midrule 
fr & fr\_gsd & 94.45 & \textbf{94.62} & 94.40 & 94.57 & 94.46 & 94.49 & 94.42 & 94.44 & 94.54 \\
 & fr\_sequoia & 88.55 & 85.85 & 87.86 & 91.32 & 90.91 & 86.17 & 86.46 & 91.68 & \textbf{91.73} \\
 & fr\_spoken & 79.15 & 83.80 & 83.33 & 84.30 & 83.89 & 83.47 & 83.20 & 84.09 & \textbf{84.50} \\
\midrule 
indic & hi\_hdtb & 92.54 & 92.47 & 92.56 & \textbf{92.66} & 92.37 & 92.49 & 92.37 & 92.51 & 92.34 \\
 & ur\_udtb & 81.02 & 81.78 & 81.93 & \textbf{82.01} & 81.74 & 81.66 & 81.41 & 81.77 & 81.42 \\
\midrule 
iranian & kmr\_mg$^*$ & 21.43 & 14.29 & 16.67 & \textbf{33.33} & 30.95 & 21.43 & 19.05 & 28.57 & 14.29 \\
 & fa\_seraji & \textbf{87.01} & 86.96 & 86.84 & 86.37 & 86.28 & 86.32 & 86.30 & 86.54 & 86.27 \\
\midrule 
it & it\_isdt & 92.92 & 93.33 & 93.22 & 92.96 & 93.09 & \textbf{93.44} & 93.26 & 93.24 & 93.29 \\
 & it\_postwita & 79.66 & 80.00 & 80.34 & 80.34 & 80.26 & 79.96 & 80.08 & \textbf{80.72} & 80.16 \\
\midrule 
ko & ko\_gsd & \textbf{82.80} & 71.46 & 79.63 & 82.71 & 82.24 & 69.41 & 79.13 & 82.55 & 82.77 \\
 & ko\_kaist & 87.05 & 81.52 & 86.52 & \textbf{87.66} & 87.54 & 81.12 & 85.81 & 87.07 & 86.86 \\
\midrule 
n-ger & da\_ddt & 86.24 & 86.37 & 86.32 & \textbf{87.05} & 86.50 & 85.90 & 85.30 & 86.41 & 86.31 \\
 & no\_bokmaal & 93.91 & 94.18 & \textbf{94.77} & 94.31 & 94.26 & 94.17 & 94.28 & 94.15 & 94.30 \\
 & no\_nynorsk & 92.57 & 92.85 & 93.21 & \textbf{93.30} & 92.93 & 92.73 & 92.47 & 92.93 & 93.15 \\
 & no\_nynorsklia & 74.23 & 76.72 & 77.18 & 76.94 & 77.26 & 76.76 & 76.85 & \textbf{77.39} & 77.18 \\
 & sv\_lines & 84.93 & 85.07 & 85.43 & 86.03 & 86.05 & 85.15 & 85.30 & \textbf{86.19} & \textbf{86.19} \\
 & sv\_talbanken & 83.85 & 84.30 & 85.32 & 85.85 & 85.90 & 85.18 & 85.41 & \textbf{86.64} & 86.14 \\
\midrule 
old & grc\_proiel & 75.57 & 77.37 & 77.26 & \textbf{77.46} & 76.91 & 76.76 & 76.70 & 76.30 & 76.47 \\
 & grc\_perseus$^*$ & 60.48 & \textbf{65.13} & 64.86 & 64.17 & 64.35 & 64.52 & 64.41 & 63.93 & 64.29 \\
 & got\_proiel$^*$ & 72.74 & 80.36 & \textbf{80.75} & 79.87 & 79.91 & 78.97 & 78.72 & 78.72 & 79.50 \\
 & la\_ittb & 90.08 & 89.98 & 90.17 & 89.93 & 89.62 & \textbf{90.30} & 90.13 & 90.08 & 90.16 \\
 & la\_proiel & 77.59 & 79.56 & 79.76 & \textbf{79.99} & 78.84 & 79.07 & 78.96 & 79.41 & 79.13 \\
 & la\_perseus & 56.28 & 65.34 & 65.63 & 70.07 & 69.72 & 64.71 & 65.40 & \textbf{71.56} & 70.05 \\
 & cu\_proiel$^*$ & 61.24 & 64.48 & \textbf{65.22} & 64.88 & 63.91 & 64.11 & 64.42 & 64.17 & 64.93 \\
\midrule 
pt-gl & gl\_ctg & 81.16 & 80.92 & 80.94 & \textbf{81.70} & 81.38 & 80.93 & 80.89 & 81.50 & 81.56 \\
 & gl\_treegal & 70.80 & 65.73 & 75.01 & 81.07 & 82.11 & 64.60 & 68.93 & \textbf{82.49} & 82.28 \\
 & pt\_bosque & 90.45 & 90.41 & 90.43 & 90.52 & \textbf{90.62} & 90.49 & 90.28 & 90.44 & 90.09 \\
\midrule 
sw-sla & hr\_set & 88.20 & 88.33 & 88.43 & 88.39 & 88.17 & 88.35 & 88.73 & \textbf{88.84} & 88.74 \\
 & sr\_set & 87.20 & 87.78 & 88.89 & 88.94 & 88.88 & 88.31 & 89.16 & 89.47 & \textbf{89.48} \\
 & sl\_ssj & 93.22 & \textbf{93.54} & 93.29 & 93.39 & 93.44 & 93.37 & 93.30 & 93.40 & 93.29 \\
 & sl\_sst & 60.30 & 70.67 & 70.78 & 70.40 & 70.28 & 70.31 & 70.84 & \textbf{71.22} & 71.09 \\
\midrule 
turkic & bxr\_bdt$^*$ & 19.51 & 26.83 & 31.71 & \textbf{36.59} & 21.95 & 26.83 & 21.95 & 29.27 & 31.71 \\
 & kk\_ktb & 14.02 & \textbf{62.62} & 58.88 & 48.60 & 49.53 & 59.81 & 61.68 & 51.40 & 49.53 \\
 & tr\_imst & 65.57 & 66.02 & 65.66 & 64.48 & 64.86 & \textbf{66.29} & 66.03 & 65.99 & 66.00 \\
 & ug\_udt$^*$ & 47.85 & 49.11 & 49.01 & 49.18 & 48.68 & \textbf{50.39} & 49.86 & 50.06 & 49.91 \\
\midrule 
w-sla & cs\_cac & 91.86 & 92.24 & 92.19 & 92.20 & 92.18 & 92.12 & 92.22 & 92.09 & \textbf{92.40} \\
 & cs\_fictree & 92.97 & 94.01 & 94.23 & 94.23 & 94.41 & 94.11 & 94.35 & \textbf{94.52} & 94.10 \\
 & cs\_pdt & 89.57 & 90.39 & 90.57 & 90.45 & 90.42 & 90.32 & 90.34 & 90.89 & \textbf{90.89} \\
 & pl\_lfg & 95.41 & 93.30 & 96.17 & \textbf{96.49} & 96.40 & 92.87 & 96.19 & 96.43 & 96.43 \\
 & pl\_pdb & 91.64 & 91.29 & 91.82 & 91.72 & 91.72 & 91.74 & 91.85 & 91.72 & \textbf{91.94} \\
 & sk\_snk & 92.07 & 93.62 & 93.51 & \textbf{93.77} & 93.58 & 93.13 & 93.41 & 93.29 & 93.02 \\
 & hsb\_ufal$^*$ & 14.47 & 59.21 & 60.53 & 63.16 & 59.21 & 61.84 & \textbf{65.79} & 61.84 & 63.16 \\
\midrule
  avg. &  &78.52 & 80.63 & 81.58 & \textbf{82.16} & 81.77 & 80.60 & 81.26 & \textbf{82.10} & 81.88\\
\bottomrule
\end{tabular}}
\caption{LAS scores for each dataset (dev) for all of our settings, both when training a parser per cluster (``Trained on cluster''), as well as having one parser for all treebanks (``Trained on all''). Bold: highest score for this training setup, omitted if the \textsc{mono} baseline performs best. $^*$ not used in mBERT pretraining.}
\label{tab:results}
\end{table}

Full results can be found in Table~\ref{tab:results}. We can see that dataset embeddings still seem to be largely useful, outperforming the monolingual baseline (\textsc{mono}) in almost all cases (56/59 test sets) and the concatenated baselines (\textsc{concat}) in a majority of cases (40/59). The dataset embedding methods are on average 3 LAS points above the \textsc{mono} baseline. These gains are even larger than the gains observed in \citet{smith-etal-2018-82} which indicates that dataset embeddings are still relevant when using large pretrained multilingual LMs. It should be noted though, that a large portion of this gain is already present when using the \textsc{concat} strategy. \textsc{Encoder} scores highest overall, but for some clusters, \textsc{decoder} is competetive (af-de-nl, es-ca, it, old, sw-sla). These clusters have in common that they are small and/or contain relatively high-resource languages (which are probably better represented in the mBERT embeddings).

For treebanks from languages not used during mBERT pretraining, scores are very low for the \textsc{mono} baseline, but they gain a lot from training on other languages. Kazakh also has very low scores, this is probably because the training split is very small. It gains a lot already in \textsc{concat}, likely because Turkish is a related language, but then loses accuracy when using dataset embeddings compared to \textsc{concat}, likely because there is not enough in-language data to learn an accurate dataset embedding.

\subsection{Results on Subsets}

\begin{figure*}
    \includegraphics[width=\textwidth]{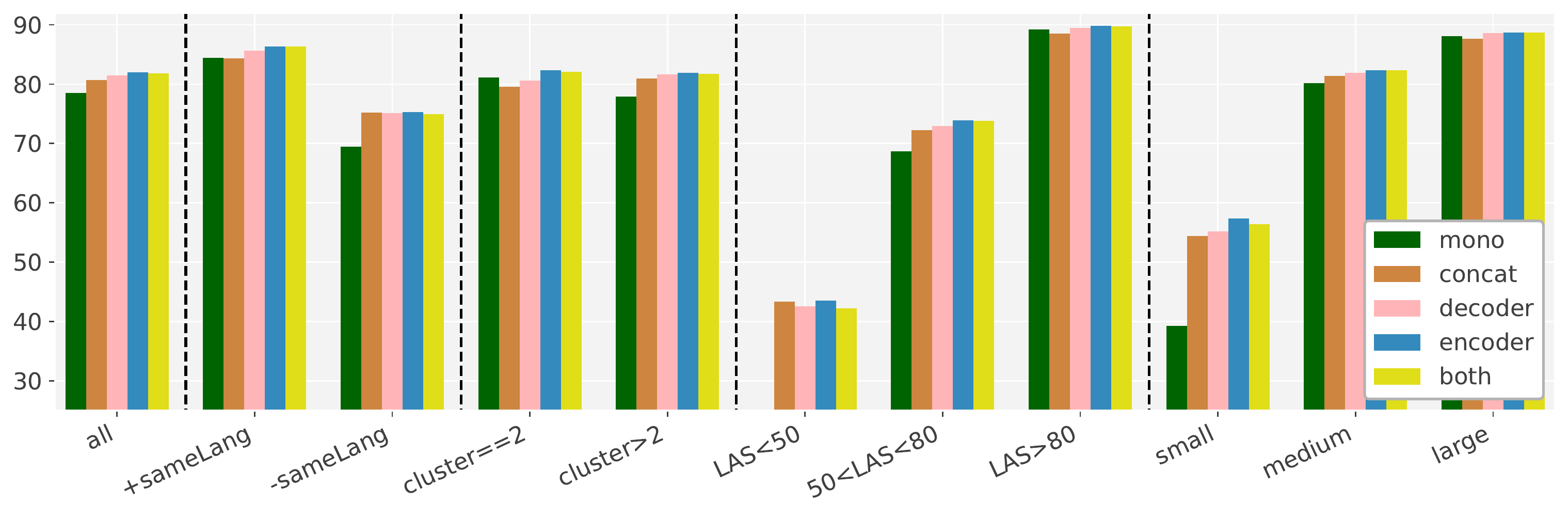}
    \caption{Average LAS scores (dev) over different subsets of the data. \texttt{All}: all data, \texttt{+sameLang}: datasets for which another in-language treebank exists, \texttt{cluster==2}: clusters of size 2, \texttt{LAS<50}: treebanks for which the `mono´ baseline scores <50 LAS, \texttt{small, medium, large}: datasets with a maximum size of respectively: 1,000, 10,000, 20,000 sentences.}
    \label{fig:filters}
\end{figure*}

To find trends in our results, we report scores over different subsets of the data. In Figure~\ref{fig:filters}, we report the results when training a parser for each cluster. The leftmost part of the graph shows the scores averaged over all datasets. \texttt{+sameLang} are the average scores for all datasets for which a dataset in the same language is included in the cluster, and \texttt{-sameLang} for all datasets for which this is not the case. \texttt{cluster==2}, are scores for clusters consisting of 2 datasets, and \texttt{cluster>2} for larger clusters. We also divide the datasets based on their performance with the \textsc{mono} baseline (LAS<50, 50<LAS<80, LAS>80), and finally based on the size of the training data: small (<1,000 sentences), medium (<10,000) and large (>20,000).

Dataset embeddings are especially beneficial for datasets where performance of the \textsc{mono} baseline is low (\texttt{LAS<50}) and for small datasets. They help moderately for medium sized datasets, and for datasets where performance of \textsc{mono} is mediocre (\texttt{50<LAS<80}). For larger and high-performing datasets, performance increases diminish. The \textsc{concat} baseline outperforms the \textsc{mono} baseline in most setups, except for small clusters indicating that some of the gains from using the dataset embedding methods in these settings are due to the additional use of data.

Perhaps a bit counterintuively, the dataset embeddings methods improve results more for treebanks for which there is not a treebank of the same language (\texttt{-sameLang}). However, this may very well be due to a confounding factor: the scores are generally a lot higher in the \texttt{+sameLang} setting than in the \texttt{-sameLang} setting and the method seems to work better for treebanks for which the baseline scores are lower.

Overall, the \textsc{encoder} model performs best; it either outperforms all others, or performs on par with the best setup. The \textsc{encoder} model outperforms the other models mostly on datasets where the \textsc{mono} baseline is low, and for small datasets. The \textsc{decoder} strategy is only beneficial in some of the data subsets, and should be used with caution. Perhaps surprisingly, the \textsc{both} strategy is not beneficial over the \textsc{encoder}, indicating that both strategies encode dataset information differently.

\begin{figure*}
    \includegraphics[width=\textwidth]{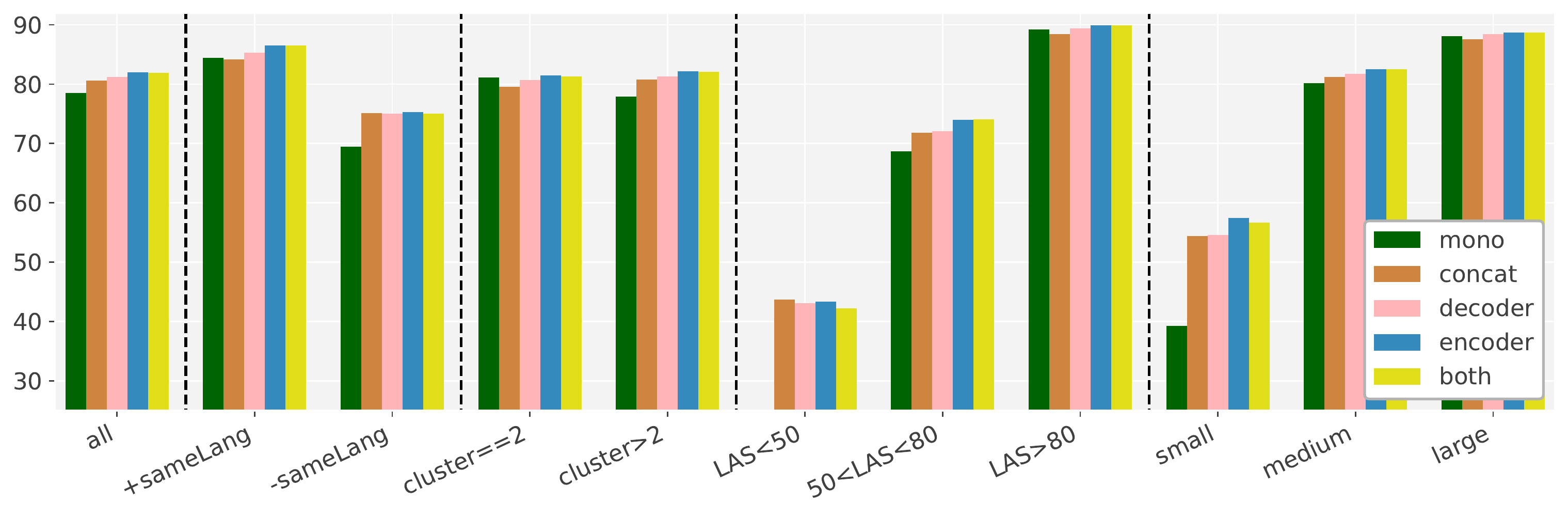}
    \caption{Scores for same subsets as Figure~\ref{fig:filters}, but when training a single parser on all data at once, instead of separate parsers for each cluster.}
    \label{fig:filtersAll}
\end{figure*}

In Figure~\ref{fig:filtersAll}, we report the results when training one parser on all the treebanks for each setup. Results are very similar to results with the parsers trained per cluster (Figure~\ref{fig:filters}). The main difference can be observed for the datasets with low performance of the \textsc{mono} baseline (LAS<50), where the difference between the different dataset embeddings and \textsc{concat} is smaller. 
The similarity to results obtained with smaller clusters indicates that 1) using dataset embeddings is also viable in a highly multilingual model and 2) a highly multilingual model might be able to pick up on dataset similarities and use the relevant data for individual languages. This has practical implications: it can be more practical to have one model that works on many languages~\citep{kondratyuk-straka-2019-75} than multiple models and it removes the need to carefully construct clusters of related languages, which can be time-consuming without a guarantee for optimal clusters. 

\subsection{Test data}
\label{sec:testData}
To avoid overusing the test data, we only confirm our main findings on the test data; we compare our best baseline to our best dataset embedding setup for both the setup trained in clusters and when training on all datasets simultaneously. Altough~\citet{van-der-goot-2021-need} suggests to concatenate the tune set to the training data for the final comparison on test, we only use the training split to save compute and because we are not aiming for a new state-of-the-art.

Results (Table~\ref{tab:test}) show that for parsers trained on clusters the \textsc{concat} baseline performs a bit higher compared to the development data (Table~\ref{tab:results}), and the gain is thus smaller, but still substantial. When training on all datasets, the results are similar. Overall, the results on the test data confirm the findings on the development data: 1) dataset embeddings are useful in both setups, 2) one large multilingual model trained on all treebanks performs on par with multiple parsers trained on clusters of related treebanks.

\begin{table}
    \centering
    \begin{tabular}{r r | r r}
        \toprule
        \multicolumn{2}{c|}{Clusters} & \multicolumn{2}{c}{All} \\
        \midrule
        \textsc{concat} & \textsc{encoder} & \textsc{concat} & \textsc{encoder} \\
        \midrule
         81.08 & \textbf{82.05} & 80.57 & \textbf{82.28} \\
         \bottomrule
    \end{tabular}
    \caption{Average LAS scores on the test data from our best baseline (\textsc{concat}), and the best setup with dataset embeddings (\textsc{encoder}) for both the cluster trained-parser and the parser trained on all data.}
    \label{tab:test}
\end{table}

\section{Conclusion}
We evaluated the usefulness of dataset embeddings for multilingual parsing using large pretrained multilingual LMs and found them to be useful in this context. Using this method improves over both a monolingual baseline and a baseline where training treebanks are concatenated, and across many settings. This method helps mostly for small treebanks.
Using dataset embeddings in the encoder showed overall slightly better results than our other embedding strategies, even better than combining the two approaches to use dataset embeddings. 
Finally, we found that using dataset embeddings in a multilingual parser that uses training data from all available treebanks works just as well as using them with clusters of treebanks from related languages. 

\section*{Acknowledgements}
We would like to thank the anonymous reviewers, Ahmet {\"U}st{\"u}n, Max M{\"u}ller-Eberstein and Daniel Varab for their discussions about dataset embeddings and evaluation. Miryam de Lhoneux was funded by the Swedish Research Council (Grant 2020-00437).

%
\blfootnote{
    %
    %
    %
    %
    %
    %
}


\bibliographystyle{acl_natbib}
\bibliography{main}

\clearpage
\appendix

\section{Exact Scores}
\begin{table}[h!]
\resizebox{\textwidth}{!}{
\begin{tabular}{l r r r r r r r r r}
\toprule
filter & mono & concat & decoder & encoder & both & ALLconc. & ALLdec. & ALLenc. & ALLboth \\
\midrule
Avg.-58 & 78.50 & 80.71 & 81.45 & 81.96 & 81.82 & 80.57 & 81.20 & \textbf{82.03} & 81.94 \\
\midrule
hasSameLang-35 & 84.44 & 84.36 & 85.62 & 86.36 & 86.34 & 84.17 & 85.26 & \textbf{86.48} & 86.48 \\
noSameLang-23 & 69.46 & 75.17 & 75.11 & 75.26 & 74.95 & 75.09 & 75.03 & \textbf{75.26} & 75.03 \\
\midrule
cluster=2-10 & 81.15 & 79.55 & 80.56 & \textbf{82.31} & 82.11 & 79.52 & 80.71 & 81.51 & 81.33 \\
cluster>2-48 & 77.94 & 80.96 & 81.64 & 81.89 & 81.76 & 80.79 & 81.30 & \textbf{82.14} & 82.07 \\
\midrule
low-5 & 22.66 & 43.36 & 42.57 & 43.54 & 42.23 & \textbf{43.69} & 43.14 & 43.38 & 42.20 \\
medium-14 & 68.68 & 72.26 & 72.97 & 73.87 & 73.84 & 71.77 & 72.04 & 74.00 & \textbf{74.04} \\
high-39 & 89.18 & 88.54 & 89.48 & 89.79 & 89.77 & 88.46 & 89.37 & 89.87 & \textbf{89.87} \\
\midrule
small-7 & 39.26 & 54.45 & 55.15 & 57.35 & 56.45 & 54.37 & 54.55 & \textbf{57.48} & 56.71 \\
medium-27 & 80.15 & 81.37 & 81.90 & 82.36 & 82.32 & 81.18 & 81.70 & 82.49 & \textbf{82.50} \\
large-24 & 88.09 & 87.64 & 88.62 & \textbf{88.68} & 88.66 & 87.52 & 88.42 & 88.67 & 88.67 \\
\bottomrule
\end{tabular}}
\caption{Exact numbers for results in Figure~\ref{fig:filters}. Average LAS scores over different subsets of the data. All: all data, +sameLang: datasets for which another in-language treebank exists, cluster==2: clusters of size 2, LAS<50: treebanks for which the `mono´ baseline scores <50 LAS, small, medium, large: datasets with a maximum size of respectively: 1,000, 10,000, 20,000 sentences.}
\label{tab:filters}
\end{table}


\end{document}